\def\year{2019}\relax
\documentclass[letterpaper]{article} 
\usepackage{aaai19}  
\usepackage{times}  
\usepackage{helvet}  
\usepackage{courier}  
\usepackage{url}  
\usepackage{graphicx}  
\usepackage{amssymb}
\usepackage{amsmath}
\usepackage{pbox}
\usepackage{booktabs}
\usepackage{multirow}

\begin{document}
%
\title{Point Cloud Processing via Recurrent Set Encoding}
\author{Pengxiang Wu$ ^1 $, Chao Chen$ ^2 $, Jingru Yi$ ^1 $, Dimitris Metaxas$ ^1 $ \\
	$ ^1 $Department of Computer Science, Rutgers University, NJ, USA, \{pw241, jy486, dnm\}@cs.rutgers.edu\\
	$ ^2 $Department of Biomedical Informatics, Stony Brook University, NY, USA, chao.chen.cchen@gmail.com
}
\maketitle
\begin{abstract}
	We present a new permutation-invariant network for 3D point cloud processing. Our network is composed of a recurrent set encoder and a convolutional feature aggregator. Given an unordered point set, the encoder firstly partitions its ambient space into parallel beams. Points within each beam are then modeled as a sequence and encoded into subregional geometric features by a shared recurrent neural network (RNN).
	The spatial layout of the beams is regular, and this allows the beam features to be further fed into an efficient 2D convolutional neural network (CNN) for hierarchical feature aggregation. Our network is effective at spatial feature learning, and competes favorably with the state-of-the-arts (SOTAs) on a number of benchmarks. Meanwhile, it is significantly more efficient compared to the SOTAs.
\end{abstract}

\section{Introduction}
\label{sec:intro}

Point cloud is a simple and compact geometric representation of 3D objects, and has been broadly used as the standard output of various sensors. In recent years, the analysis of point clouds has gained much attention due to its wide application in real world problems such as autonomous driving \cite{CVPR_autonomous_driving}, robotics \cite{Robotics}, and navigation \cite{Navigation}.
However, it is nontrivial to solve such tasks using traditional deep learning tools, e.g., convolutional neural networks (CNNs).
Unlike a 2D image with regularly packed pixels, a point cloud consists of sparse points without a canonical order. Moreover, the spatial distribution of a point cloud is heterogeneous due to factors in data acquisition, e.g., perspective effects and radial density variations.

Due to the 3D nature of the problem, various methods have been proposed to convert a point cloud into a 3D volumetric representation, to which 3D CNNs are then applied \cite{3D_ShapeNets,VoxNet}.
However, despite their success in analyzing 2D images, CNNs are not satisfactory in this context. 
The commonly used 3D CNN is extremely memory consuming, and thus can not be trained efficiently.
A more serious issue is that converting a point cloud into a volumetric representation introduces quantization artifacts and loses fine-scale
geometric details.

\begin{figure*}[t!]
	\centering
	\includegraphics[width=\linewidth]{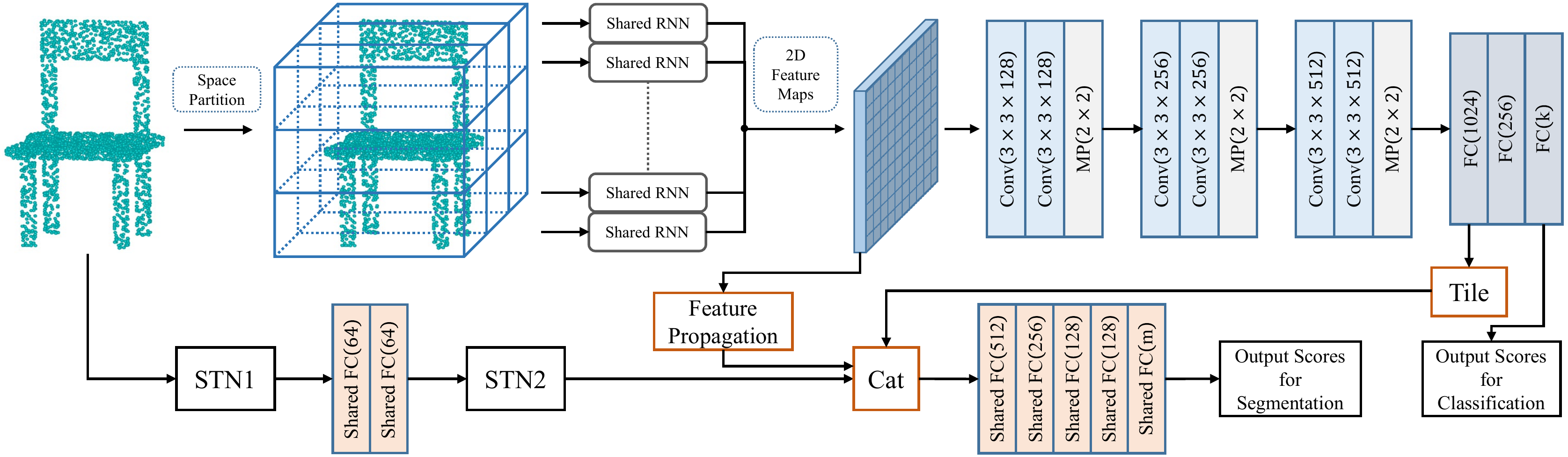}
	\caption{The architecture of RCNet. In the recurrent set encoder, the ambient space of input points is partitioned into parallel beams, where the enclosed points are encoded by a shared RNN. The subregional features from each beam are later processed by a 2D CNN. Depending on the tasks, the aggregated global features are fed forward directly for shape prediction, or tiled and concatenated with the per-point features for semantic segmentation. The feature propagation refers to the operation that propagates the non-local features within each beam to the corresponding component points. The other operations used are: Conv (2D convolution), MP (2D max-pooling), FC (fully connected layer). Batchnorm and ReLU are
		used in all layers except the last one, and the shared FC is applied per point. Numbers in parentheses represent the size of operation, and the hidden size of RNN is 64 and 128 for classification and segmentation tasks, respectively. The STN block refers to the spatial transformer network \cite{STN,PointNet}. It outputs a transformation matrix and is comprised of a shared MLP(64, 128, 1024), a global max-pooling and another MLP(512, 256, $ d^2 $), where $ d $ is the number of features per input point.
	}
	\label{fig:architecture}
\end{figure*}

Better performance has been achieved by deep networks that avoid the volumetric convolutional architechture and operate directly on point clouds. Representative works include PointNet \cite{PointNet} and PointNet++ \cite{PointNet++}, which process point clouds by combining multi-layer perceptron (MLP) network with symmetric operations (e.g., max-pooling) to learn point features globally or hierarchically. 
Inspired by PointNet, several recent methods have been proposed to further improve the point feature representation \cite{Kernel-Graph,ShapeContextNet,SO-Net}. This class of networks are invariant to input permutation and have achieved state-of-the-art results. However, due to the reliance on the coarse feature pooling technique, they fail to fully exploit fine-scale geometric details.

In this work, we aim to completely bypass the coarse pooling-based technique, and propose a new deep network for point cloud data. At the core of our method is a \textit{recurrent set encoder}, which divides the ambient domain into parallel \textit{beams} and encodes the points within each beam as \emph{subregional} geometric features with an RNN. Our key observation is that when the beam is of moderate size, the RNN is approximately dealing with a sequence of points, as a beam only contains points near a 1D line. Such a sequential input largely benefits the learning of RNN. 
Meanwhile, noticing that the beams are packed in a regular spatial layout, we use a 2D CNN to further analyze the beam features (called the \emph{convolutional feature aggregator}). Being efficient and powerful at feature learning, the 2D CNN can effectively aggregate the subregional features into a global one, while further benefiting the RNN learning in return. Our method (see Fig.~\ref{fig:architecture}) is surprisingly efficient and effective for point cloud processing. It is invariant to point order permutation, and competes favorably with the state-of-the-arts (SOTAs) in terms of both accuracy and computational efficiency. 

A few recent works also adopt convolution for point cloud processing. They typically utilize carefully designed domain transformations to map point data into suitable spaces, where convolution could be applied. Examples include SPLATNet \cite{SPLATNet} and PCNN \cite{PCNN_2018}. However, these methods are inefficient as they rely on sophisticated geometric transformations and complex convolutional operations, e.g., continuous volumetric convolution or sparse bilateral convolution. In contrast, our method only employs regular spatial partitioning and sorting, and leverages classic neural network architectures such as RNN and 2D CNN, which are well supported at both software and hardware levels. As a result, our network circumvents much implementation overhead and is significantly more efficient than these SOTAs in computation. It is worth mentioning that, our recurrent set encoder can be seen as a domain mapping function as well. But unlike these SOTAs, it is automatically learned via back-propagation instead of by careful handcrafted design.

In this work, we focus on point cloud classification and segmentation tasks, and evaluate the proposed method on several datasets, including ModelNet10/40 \cite{3D_ShapeNets}, ShapeNet part segmentation \cite{ShapePartSeg}, and S3DIS \cite{S3DIS}. Experimental results demonstrate the superior performance of our method to the SOTAs in both accuracy and computational efficiency. 

In a nutshell, our main contributions are as follows:

\begin{itemize}
	\item We present a new architecture that operates directly on point clouds without relying on symmetric functions (e.g., max-pooling) to achieve permutation invariance.
	\item We propose a recurrent set encoder for effective subregional feature extraction. To the best of our knowledge, this is the first time an RNN is effectively employed to model point clouds directly.
	\item We propose to introduce the 2D CNN for aggregating subregional features. This design maximally utilizes the strengths of CNN while further benefiting the RNN encoder. The resulting network is efficient as well as effective at hierarchical and spatially-aware feature learning.
\end{itemize}


\section{Related Work}
\label{sec:related_work}

We briefly review the existing deep learning approaches for 3D shape processing, with a focus on point cloud setting.

\subsubsection{Volumetric Methods} One classical approach to handling unstructured point clouds or meshes is to first rasterize them into regular voxel grids, and then apply standard 3D CNNs \cite{3D_ShapeNets,VoxNet,3D_CNN_LiDAR,Vol_MVCNN,Orientation_VoxelNet,Segcloud,3DCNN-DQN-RNN}. The major issue with such volumetric representations is that they tend to produce sparsely-occupied grids, which are unnecessarily memory-consuming. Besides, the grid resolutions are limited due to excessive memory and computational cost, causing quantization artifacts and loss of details. To remedy these issues, recent methods propose to adaptively partition the grids and place denser cells near the shape surface \cite{O-CNN,OctNet,Octree}. These methods suffer less from the computational and memory overhead, but still lose geometric details due to sampling and discretization.

\subsubsection{View-based Methods} Another strategy is to encode the 3D shapes via a collection of 2D images which are rendered from different views. These rendered images can be fed into traditional 2D CNNs and processed via transfer learning, i.e., fine-tuning networks pre-trained on large-scale image datasets \cite{MVCNN,Vol_MVCNN,PCNN}. However, such view projections would lead to self-occlusions and consequently severe loss of geometric information. Moreover, view-based methods are mostly applied to classification tasks, and are hard to generalize to detail-focused tasks such as shape segmentation and completion.

\subsubsection{Non-Euclidean Methods} These approaches build graphs from the input data (e.g., based on the mesh connectivity or k-nearest neighbor relationship), and apply CNNs to the graph spectral domain for shape feature learning \cite{GeoDL,LocalizedSpec_CNN,GraphCNN,FastGraphCNN,SemiSupervised_GraphCNN,AdapGraphCNN}. Graph CNN models are suitable for non-rigid shape analysis due to the isometry invariance. However, it is comparatively difficult to generalize these methods across non-isometric shapes with different structures, largely because the spectral bases are domain-dependent \cite{SyncSpecCnn}.

\subsubsection{Point Cloud-based Methods} PointNet \cite{PointNet} pioneers a new type of deep neural networks that act directly on point clouds without data conversions. Its key idea is to learn per-point features independently, and then aggregate them in a permutation-invariant manner via a symmetric function, e.g., max-pooling. While achieving impressive performance, PointNet fails to capture crucial fine-scale structure details. To address this issue, the follow-up work PointNet++ \cite{PointNet++} exploits local geometric information by hierarchically stacking PointNets. This leads to improved performance, but at the cost of computational efficiency. Besides, since PointNet++ still treats points individually at local scale, the relationships among points are not fully captured. In light of the above challenges, a number of recent works have been proposed for better shape modeling \cite{KD_net,SO-Net,Kernel-Graph,SliceNet,ShapeContextNet,PCCN}. These methods overcome the weakness of coarse pooling operation at some degree, and achieve improved performance.

Another class of methods have been recently developed without relying on pooling to guarantee permutation invariance. They typically transform the point data into another domain, where convolutions could be readily applied. In SPLATNet \cite{SPLATNet}, the source point samples are mapped into a high-dimensional lattice, where sparse bilateral convolution is employed for shape feature learning. In PCNN \cite{PCNN_2018}, a pair of extension and restriction operators are designed to translate between point clouds and volumetric functions, such that continuous volumetric convolution could be applied. Our method could be considered belonging to this category from the perspective of domain transformation. However, different from existing methods, our domain mapping function is automatically learned rather than by handcrafted design. Moreover, instead of utilizing complex convolutions, we employ the classic 2D convolution for feature aggregation. As a result, our method is more efficient in computation as well as effective at point feature learning.


\section{Method}
\label{sec:method}

In this work, we focus on two tasks: point cloud classification and segmentation, and present two architectures correspondingly, as illustrated in Fig.~\ref{fig:architecture}. The input is a point set $ P = \{p_i \in \mathbb{R}^d, i=1, \cdots, N \} $, where each point $ p_i $ is a vector of coordinates plus additional features, such as normal and color. The output will be a $ 1\times K $ score vector for classification with $ K $ classes, or an $ N \times M $ score matrix for segmentation with $ M $ semantic labels. Our network, termed RCNet, consists of two components: the \textit{recurrent set encoder} and the \textit{convolutional feature aggregator}. The recurrent set encoder aims to extract subregional features from input point cloud, while convolutional feature aggregator is responsible for aggregating these extracted features hierarchically. Below we explain their details.

\subsubsection{Recurrent Set Encoder} Given an unordered point set, the recurrent set encoder firstly partitions the ambient space into a set of parallel beams, and then divides the points into subgroups accordingly (see Fig.~\ref{fig:architecture}). The beams are uniformly distributed in a structured manner, spanning a 2D lattice. In particular, suppose the width, height and depth of a beam extends along $ x $, $ y $ and $ z $ axis, respectively. Let $ r $ and $ s $ be the hyper-parameters controlling the number of beams: $ w=(x_{max} - x_{min})/r $ and $ h=(y_{max} - y_{min})/s $, where $ w,h $ are the beam width and height; $ [x_{min}, x_{max}] $ and $ [y_{min}, y_{max}] $ are the maximum spanning ranges of points. Then a point with coordinate $ (x_k, y_k, z_k) $ is assigned to the $ (i,j) $-th beam if $ x_k - x_{min} \in [(i-1)w, iw) $ and $ y_k - y_{min} \in [(j-1)h, jh) $. In our implementation, since the point clouds are normalized to fit within a unit ball, we can simply set $ x_{min} = y_{min}=-1 $ and $ x_{max} = y_{max}=1 $. The subgroups of points are denoted by $ \{S_{ij}\}_{i=1, j=1}^{r,s} $. Note that depending on the tasks, it is also possible to perform non-uniform partition \cite{O-CNN}. In this work we only focus on uniformly partitioned beams.

Given points in subgroup $ S_{ij} $, we treat them as a sequential signal and process it with an RNN. In particular, before being fed to RNN, points within each beam are sorted along the beam depth (according to their $ z $ coordinates). The RNN is single-directional, implemented using Gated Recurrent Units (GRU) \cite{GRU} with 2 layers. 
To the best of our knowledge, our network is the first to \textit{effectively} use an RNN to handle 3D point sets directly.
Interestingly, it has been previously observed that an RNN performs poorly on a 3D point cloud due to the lack of a unique and stable ordering \cite{PointNet,OrderMatters}.
The key to our success is the beam partition strategy.  
With the relatively dense partitioning, the points within each beam is of moderate size, and can be approximately considered distributed along a 1D line. In another word, the RNN is approximately handling point signal of moderate length in a 1D space. This facilitates the learning of RNN and makes it behave quite robustly with respect to the input perturbation. 

The output of recurrent set encoder is a grid of 1D feature vectors, which are taken as a 2D feature map and fed into the subsequent 2D CNN aggregator:
\begin{equation}
\label{eq:feature_map}
I =
\begin{bmatrix}
\mathcal{R}(S_{11}) & \dots & \mathcal{R}(S_{1s})\\
\vdots & \ddots & \vdots \\
\mathcal{R}(S_{r1}) & \dots & \mathcal{R}(S_{rs})
\end{bmatrix},
\end{equation}
where $ \mathcal{R} $ is a shared RNN with hidden size $ \ell $, and $ I \in \mathbb{R}^{r\times s \times \ell} $. Note that, we only utilize RNN to encode nonempty beams, and for those empty ones we pad zero vectors at the corresponding positions of $ I $. 

\subsubsection{Convolutional Feature Aggregator} We first note that the features encoded by RNN are actually \textit{non-local}, as the points within each beam span a large range along the beam depth. To build a global shape descriptor, we need to connect these non-local features. A natural choice is using 2D convolutional neural network, given the structured output $ I $ in Eq.(\ref{eq:feature_map}). Being efficient and powerful at multi-scale feature learning, a 2D CNN aggregator brings much computational and modeling advantage
compared to the sophisticated aggregators in previous methods, as shown in the experiment section. Further,
the strength of
a 2D CNN alleviates the modeling burden of the recurrent encoder and boosts the overall
performance. In this work, we utilize a simple shallow CNN architecture to validate our idea (see Fig.~\ref{fig:architecture}), and leave advanced architectures for future exploration.

The aggregated global feature could be used for shape classification directly, or combined with the per-point features for semantic segmentation, as illustrated in Fig.~\ref{fig:architecture}. Note that, for segmentation task we inject additional subregional information into the points via feature propagation, so as to facilitate the discriminative point feature learning.

\subsubsection{Remarks} We stress a few key properties of RCNet below.
\begin{enumerate}
	\item It is invariant to point permutation, a result derived from point sorting within beams.
	\item The amount of context information embedded in the 2D feature maps can be controlled with beam sizes. Smaller beams would preserve richer spatial contexts while larger ones would contain less. In the extreme case, when the ambient space is trivially partitioned, i.e., there is only one beam, RCNet degenerates to the vanilla RNN model for point clouds \cite{PointNet}. The effect of beam size will be investigated in the experiment section in detail.
	\item RCNet is computationally efficient and converges fast during training, due to the benefits of 2D CNN. Besides, unlike vanilla RNN, our recurrent encoder is parallelizable with each RNN processing a small portion of points. This further facilitates the computational efficiency.
\end{enumerate}

\subsection{RCNet Ensemble}
In RCNet, the beam depth extends along a certain direction, i.e., $ z $ axis. While being effective at extracting subregional features in this direction, the recurrent encoder does not explicitly consider features along other directions. To further facilitate the point feature learning, we propose to capture geometric details in different directions and use an ensemble of RCNets, of which each single model has different beam depth directions. 
The ensemble unifies a set of ``weak'' RCNets and is able to learn richer geometric features. The resulting model, termed RCNet-E, is flexible and achieves better performance, as shown in our experiments. In practice, we implement an ensemble by independently training three RCNets, whose beam depths extend along $ x $, $ y $ and $ z $ axes respectively.
Then we simply average their predictions to produce the final results. Note that, although multiple networks are used,
thanks to the high efficiency of RCNet, their ensemble is still quite efficient. Moreover, such ensemble is amenable to parallelization for further speed-up.


\section{Experiments}
\label{sec:exp}

In this section, we evaluate our RCNet on multiple benchmark datasets, including ModelNet10/40 \cite{3D_ShapeNets}, ShapeNet part segmentation \cite{ShapePartSeg}, and S3DIS \cite{S3DIS}. In addition, we analyze the properties of RCNet in details with extensive controlled experiments. Code can be found on the authors' homepage.

\subsubsection{Ablation Study and a Baseline Model} To validate the advantages of our recurrent set encoder, we compare it with the widely used pooling-based feature aggregator. In particular, we replace the recurrent encoder in RCNet with an MLP, consisting of two layers whose sizes are the same with that of the corresponding RNN hidden layers. This MLP is shared and applied to each point, followed by a global max-pooling to aggregate the subregional features. Meanwhile, the remaining parts of the model are kept the same with RCNet. We take this modified network as a baseline model. As demonstrated in the following section, our recurrent set encoder is more effective at describing the spatial layout and geometric relationships than pooling-based technique.

\subsection{Shape Classification}

\subsubsection{Datasets} ModelNet10 and ModelNet40 \cite{3D_ShapeNets} are standard benchmarks for shape classification. ModelNet10 is composed of 3991 train and 908 test CAD models from 10 classes, while ModelNet40 consists of 12311 models from 40 categories, with 9843 models used for training and 2468 for testing. These models are originally organized with triangular meshes, and we follow the same protocol of \cite{PointNet,PointNet++} to convert them into point clouds. In particular, for each model, we uniformly sample 1024 points from the mesh, and then normalize them to fit within a unit ball, centered at the origin. We only use the point positions as input features and discard the normal information.

\subsubsection{Training} Following \cite{PointNet,PointNet++,KD_net}, we apply data augmentation during the training procedure by randomly translating and scaling the objects, as well as perturbing the point positions. We set the hyper-parameters $ r=32$ and $ s=32 $. The learning rate is initialized to 0.001 with a decay of 0.1 every 30 epochs. The networks are optimized using Adam \cite{Adam}, and it takes about $ 2 \sim 3 $ hours for the training to converge on a single NVIDIA GTX 1080 Ti GPU.

\subsubsection{Results} We compare RCNet with several state-of-the-arts: VoxNet \cite{VoxNet}, volumetric CNN \cite{Vol_MVCNN}, O-CNN \cite{O-CNN}, MVCNN \cite{MVCNN}, ECC \cite{ECC}, DeepSets \cite{Deep_Sets}, vanilla RNN and PointNet \cite{PointNet}, PointNet++ \cite{PointNet++}, KD-Net \cite{KD_net}, Pointwise CNN \cite{PointWise}, SO-Net \cite{SO-Net}, KCNet \cite{Kernel-Graph}, SCN \cite{ShapeContextNet}, and PCNN \cite{PCNN_2018}. The results are demonstrated in Table~\ref{table:classification}.

We observe that a single RCNet is able to achieve competitive results against the state-of-the-arts, and with ensemble the performance is further boosted. In particular, RCNet performs better than most existing approaches. While obtaining similar accuracy to PCNN, our network is significantly simpler in design. On the other hand, compared to the baseline model, RCNet outperforms it by a large margin. This validates the effectiveness of recurrent encoder at modeling the relative relationships among points. It is worth noting that, in \cite{SO-Net} the SO-Net also attempted to apply the standard CNN to the generated image-like feature maps, but only led to decreased performance. In contrast, our RCNet is better at incorporating the advantages of CNN into point cloud analysis, thanks to the recurrent set encoder.

\begin{table}[t!]
	\begin{center}
		\resizebox{1.0\columnwidth}{!}{
		\begin{tabular}{ccccc}
			\hline
			Method & \# Points & Input & MN10 & MN40  \\
			
			\hline
			
			VoxNet & - & Vox & 92.0 & 83.0\\
			Vol. CNN & - & Vox & - & 89.9\\
			O-CNN & - & Vox & - & 90.6\\
			MVCNN & - & Img & - & 90.1\\
			
			\hline
			
			ECC & 1000 & PC & 90.8 & 87.4 \\
			DeepSets & 5000 & PC & - & 90.0 \\
			RNN (vanilla) & 1024 & PC & - & 78.5 \\
			PointNet & 1024 & PC & - & 89.2\\
			PointNet++ & 1024 & PC & - & 90.7\\
			KD-Net & 1024 & PC & 93.3 & 90.6\\
			Pointwise CNN & - & PC & - & 86.1 \\
			SO-Net & 2048 & PC & 94.1 & 90.9\\
			KCNet & 1024 & PC & 94.4 & 91.0\\
			SCN & 1024 & PC & - & 90.0 \\
			PCNN & 1024 & PC & 94.9 & 92.3\\
			
			\hline
			
			Baseline (ours) & 1024& PC & 92.5 & 89.1 \\
			Baseline-E (ours) & 1024 & PC & 93.0 & 90.8 \\
			
			RCNet (ours) & 1024 & PC & 94.7 & 91.6 \\
			RCNet-E (ours) & 1024 & PC & \textbf{95.6} & \textbf{92.3}\\
			
			\hline
		\end{tabular}
		}
		\caption{Classification accuracies on ModelNet datasets. (``Vox": Voxels; ``Img": Images; ``PC": Point Clouds.)}
		\label{table:classification}
	\end{center}
\end{table}

Finally, our RCNet is computationally efficient. In particular, a single RCNet can be trained in about 3 hours. This is much faster than PointNet++ and PCNN, both of which require about 20 hours for training \cite{PointNet++,PCNN_2018}. Besides, as shown in Table~\ref{table:time}, on average it takes about 0.4 milliseconds for RCNet to forward a shape, while PointNet++ and PCNN require 2.8 and 16.8 milliseconds, respectively\footnote{For PCNN, we run the code released by the authors (https://github.com/matanatz/pcnn), with the default pointconv configuration. For PointNet++, we use the official implementation (https://github.com/charlesq34/pointnet2), and test the MSG model with the default network setting.}. Table~\ref{table:time} also summarizes the number of parameters of different networks. Interestingly, although our model has larger size, it still runs faster than other competitors. This validates that the classic RNN and 2D CNN, which are well supported at both software and hardware levels, contribute largely to the model efficiency. In contrast, since PointNet++ need to perform additional K-nearest neighbor query on the fly on GPU, it is much less efficient in spite of the smaller model size. Similarly, PCNN and SPLATNet$ _\text{3D} $ rely on sophisticated geometric transformations and complex convolutional operations. These operations are much less GPU-friendly and cause a lot of overhead in practice. It is worth mentioning that, since RCNet-E is naturally parallelizable, its inference time is almost the same with that of a single RCNet.

\begin{table}[t!]
	\begin{center}
		\begin{tabular}{c|cc|cc}
			\hline
			\multirow{2}{*}{Method} & \multicolumn{2}{c|}{Infer. Time (ms)} & \multicolumn{2}{c}{\# Param. (M)} \\\cline{2-5}
			& Class. & Seg. & Class. & Seg.\\
			
			\hline
			
			RCNet (ours) & \textbf{0.4} & \textbf{4.5} & 13.3 & 16.7 \\
			RCNet-E (ours) & 0.6 & 4.8 & 39.9 & 50.1 \\
			PointNet++ & 2.8 & 11.9 & \textbf{1.0} & \textbf{1.7} \\
			PCNN & 16.8 & 109.3 & 8.1 & 5.4 \\
			SPLATNet$ _{\text{3D}} $ & - & 23.1 & - & 2.7\\

			\hline 
		\end{tabular}
		\caption{Comparison of inference time and model size for different networks. Classification and segmentation are performed on ModelNet40 and ShapeNet part datasets, respectively. Time is measured in milliseconds, which correspond to the cost of forwarding a shape on average. The hardware used is an Intel i7-6850K CPU and a single NVIDIA GTX 1080 Ti GPU. ``M" stands for million.}
		\label{table:time}
	\end{center}
\end{table}

\begin{table*}[t!]
	\centering
	\begin{center}
		\resizebox{2.11\columnwidth}{!}{
		\begin{tabular}[t]{c|c| *{16}{c}}
			\hline
			
			& mean & aero & bag & cap & car & chair & ear-p & guitar & knife & lamp & laptop & motor & mug & pistol & rocket & skate & table \\
			
			\hline
			
			\# shapes &  & 2690 & 76 & 55 & 898& 3758 & 69 & 787 & 392 & 1547 & 451 & 202 & 184 & 283 & 66 & 152 & 5271 \\
			
			\hline
			
			PointNet & 83.7 & 83.4 & 78.7 & 82.5 & 74.9 & 89.6 & 73.0 & 91.5 & 85.9 & 80.8 & 95.3 & 65.2 & 93.0 & 81.2 & 57.9 & 72.8 & 80.6\\
			
			PointNet++ & 85.1 & 82.4 & 79.0 & 87.7 & 77.3 & 90.8 & 71.8 & 91.0 & 85.9 & 83.7 & 95.3 & 71.6 & 94.1 & 81.3 & 58.7 & \textbf{76.4} & 82.6\\
			
			Kd-Net & 82.3 & 80.1 & 74.6 & 74.3 & 70.3 & 88.6 & 73.5 & 90.2 & 87.2 & 81.0 & 94.9 & 57.4 & 86.7 & 78.1 & 51.8 & 69.9 & 80.3\\
			
			SPLATNet$ _{\text{3D}} $ & 84.6 & 81.9 & 83.9 & 88.6 & 79.5 & 90.1 & 73.5 & 91.3 & 84.7 & 84.5 & 96.3 & 69.7 & 95.0 & 81.7 & 59.2 & 70.4 & 81.3\\
			
			SO-Net (p.t.) & 84.9 & 82.8 & 77.8 & 88.0 & 77.3 & 90.6 & 73.5 & 90.7 & 83.9 & 82.8 & 94.8 & 69.1 & 94.2 & 80.9 & 53.1 & 72.9 & 83.0\\
			
			RSNet & 84.9 & 82.7 & \textbf{86.4} & 84.1 & 78.2 & 90.4 & 69.3 & 91.4 & 87.0 & 83.5 & 95.4 & 66.0 & 92.6 & 81.8 & 56.1 & 75.8 & 82.2\\
			
			KCNet & 84.7 & 82.8 & 81.5 & 86.4 & 77.6 & 90.3 & 76.8 & 91.0 & 87.2 & 84.5 & 95.5 & 69.2 & 94.4 & 81.6 & 60.1 & 75.2 & 81.3 \\
			
			A-SCN & 84.6 & 83.8 & 80.8 & 83.5 & 79.3 & 90.5 & 69.8 & 91.7 & 86.5 & 82.9 & 96.0 & 69.2 & 93.8 & 82.5 & \textbf{62.9} & 74.4 & 80.8\\
			
			PCNN & 85.1 & 82.4 & 80.1 & 85.5 & 79.5 & 90.8 & 73.2 & 91.3 & 86.0 & \textbf{85.0} & 95.7 & 73.2 & 94.8 & 83.3 & 51.0 & 75.0 & 81.8\\
			
			\hline
			
			Baseline (ours) & 84.6 & 83.3 & 76.8 & 87.6 & 78.6 & 90.3 & 73.7 & 90.9 & 86.8 & 82.1 & 95.5 & 69.8 & 94.3 & 82.6 & 58.4 & 76.0 & 81.7\\
			
			Baseline-E (ours) & 85.3 & 84.1 & 77.0 & 87.4 & 79.8 & 90.6 & 73.9 & 91.5 & 87.0 & 83.1 & 95.6 & 70.0 & 94.4 & 83.4 & 58.1 & 75.6 & 82.4 \\
			
			RCNet (ours) & 85.3 & 84.4 & 80.1 & 89.6 & 78.6 & 90.5 & 76.3 & 91.4 & \textbf{87.3} & 82.5 & 96.1 & 73.1 & 94.7 & 84.0 & 61.0 & 76.1 & 82.6\\
			
			RCNet-E (ours) & \textbf{86.0} & \textbf{85.3} & 81.1 & \textbf{90.0} & \textbf{79.9} & \textbf{91.1} & \textbf{77.0} & \textbf{91.8} & \textbf{87.3} & 84.1 & \textbf{96.5} & \textbf{75.1} & \textbf{95.1} & \textbf{84.8} & 61.3 & \textbf{76.4} & \textbf{83.1}\\
			
			\hline
		\end{tabular}
		}
		\caption{Results on ShapeNet part segmentation. mIoU metric is used for evaluation. The instance average mIoU as well as mIoU scores for each shape category are listed. Our RCNet-E outperforms the state-of-the-arts in most categories and achieves the best instance average mIoU.}
		\label{table:partseg}
	\end{center}
\end{table*}

\subsection{Shape Part Segmentation}

\subsubsection{Dataset and Configuration} For shape part segmentation, the task is to classify each point of a point cloud into one of the predefined part categories. We evaluate the proposed method on the challenging ShapeNet part dataset \cite{ShapePartSeg}, which contains 16881 shapes from 16 categories. The shapes are consistently aligned and normalized to fit within a unit ball. For each shape, it is annotated with 2-6 part labels, and in total there are 50 different parts. We sample 2048 points for each shape following \cite{PointNet,PointNet++}. As in \cite{PointNet++}, apart from point positions we also use normal information as input features. Following the setting in \cite{ShapePartSeg}, we evaluate our methods assuming that the category of the input 3D shape is already known. The segmentation results are reported with the standard metric mIoU \cite{PointNet}. We use the official train/test split as in \cite{shapenet2015} in our experiment. We follow the same network configuration with the classification task.

\subsubsection{Results} Table~\ref{table:partseg} compares RCNet with the following state-of-the-art point cloud-based methods: PointNet \cite{PointNet}, PointNet++ \cite{PointNet++}, Kd-Net \cite{KD_net}, SPLATNet$ _{\text{3D}} $ \cite{SPLATNet}, SO-Net (pre-trained) \cite{SO-Net}, RSNet \cite{SliceNet}, KCNet \cite{Kernel-Graph}, A-SCN \cite{ShapeContextNet}, and PCNN \cite{PCNN_2018}. In Table~\ref{table:partseg}, we report the instance average mIoU as well as the mIoU scores for each category.

\begin{figure}[t!]
	\centering
	\includegraphics[width=1.0\linewidth]{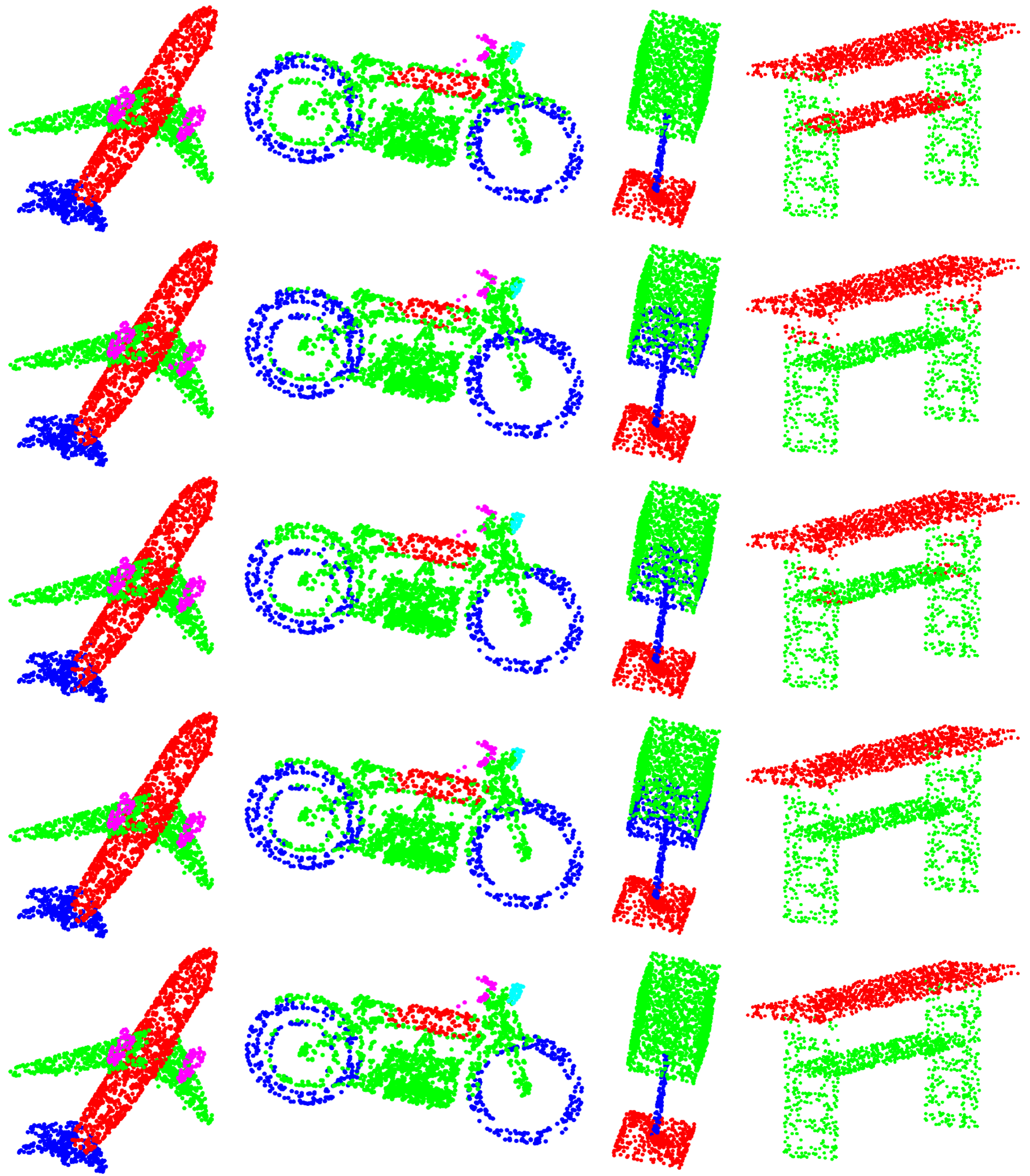}
	\caption{Visualization of ShapeNet part segmentation results. From top to bottom: ground truth, baseline, baseline-E, RCNet, RCNet-E. From left to right: airplane, motorbike, lamp, table.}
	\label{fig:shape_part_seg}
\end{figure}

As is shown, our method achieves better results than the state-of-the-art works. In particular, a single RCNet is able to achieve average mIoU of 85.3, which is competitive with the performance of PointNet++ and PCNN. With ensemble, the accuracy is further boosted and our method dominates most of the shape categories. Some qualitative segmentation results are illustrated in Fig.~\ref{fig:shape_part_seg}. Specifically, the first two columns show that both RCNet and RCNet-E are able to handle the small details of objects well. The third column indicates that the ensemble helps correct the prediction error of a single model, and is better at capturing the fine-grain semantics than the baseline methods. The last column corresponds to a failure case, which is possibly due to the imperfect model representation ability or caused by shape semantic ambiguity (i.e., the table board in the middle could be interpreted as either table support or tabletop).

In Table~\ref{table:time}, we compare the computational efficiency of different networks on part segmentation task. As is shown, our method is more efficient than the state-of-the-arts\footnote{For SPLATNet$ _{\text{3D}} $, we run the code implemented by the authors (https://github.com/NVlabs/splatnet), with the default network configuration. For PointNet++, the MSG model with one hot vector is tested. For PCNN, we use the default pointconv configuration. In the experiment we sample 2048 points for each shape.}.

\begin{table}[t!]
	\begin{center}
		\begin{tabular}{ccc}
			\hline
			Method & Mean IoU & Overall accuracy \\
			
			\hline
			PointNet & 47.71 & 78.62 \\
			A-SCN & 52.72 & 81.59 \\
			Pointwise CNN & - & 81.50\\
			\hline
			
			Baseline (ours) & 50.31 &  81.57 \\
			Baseline-E (ours) & 52.38  & 82.98 \\
			
			RCNet (ours) & 51.40 & 82.01 \\
			RCNet-E (ours) & \textbf{53.21} & \textbf{83.58} \\
			
			\hline
		\end{tabular}
		\caption{Segmentation results on S3DIS dataset. Mean IoU and point-wise accuracy are listed.}
		\label{table:S3DIS}
	\end{center}
\end{table}

\subsection{Semantic Scene Segmentation}

\subsubsection{Dataset and Configuration} We evaluate our RCNet on the scene parsing task with Standford 3D indoor scene dataset (S3DIS) \cite{S3DIS}. S3DIS consists of 6 scanned large-scale areas, which in total have 271 rooms. Each point in the scene point cloud is annotated with one of the 13 semantic categories. Following \cite{PointNet}, we pre-process the data by splitting the scene points into rooms, and then subdividing the rooms into small blocks with area 1m by 1m (measured on the floor). As in \cite{PointNet}, we also use k-fold strategy for training and testing. At training time we randomly sample 2048 points for each block, but use all the points during testing. We represent each point using 9 attributes, including XYZ coordinates, RGB values and normalized coordinates as to the room. The same shape segmentation RCNet is used for this task.

\begin{figure}[t!]
	\centering
	\includegraphics[width=1.0\linewidth]{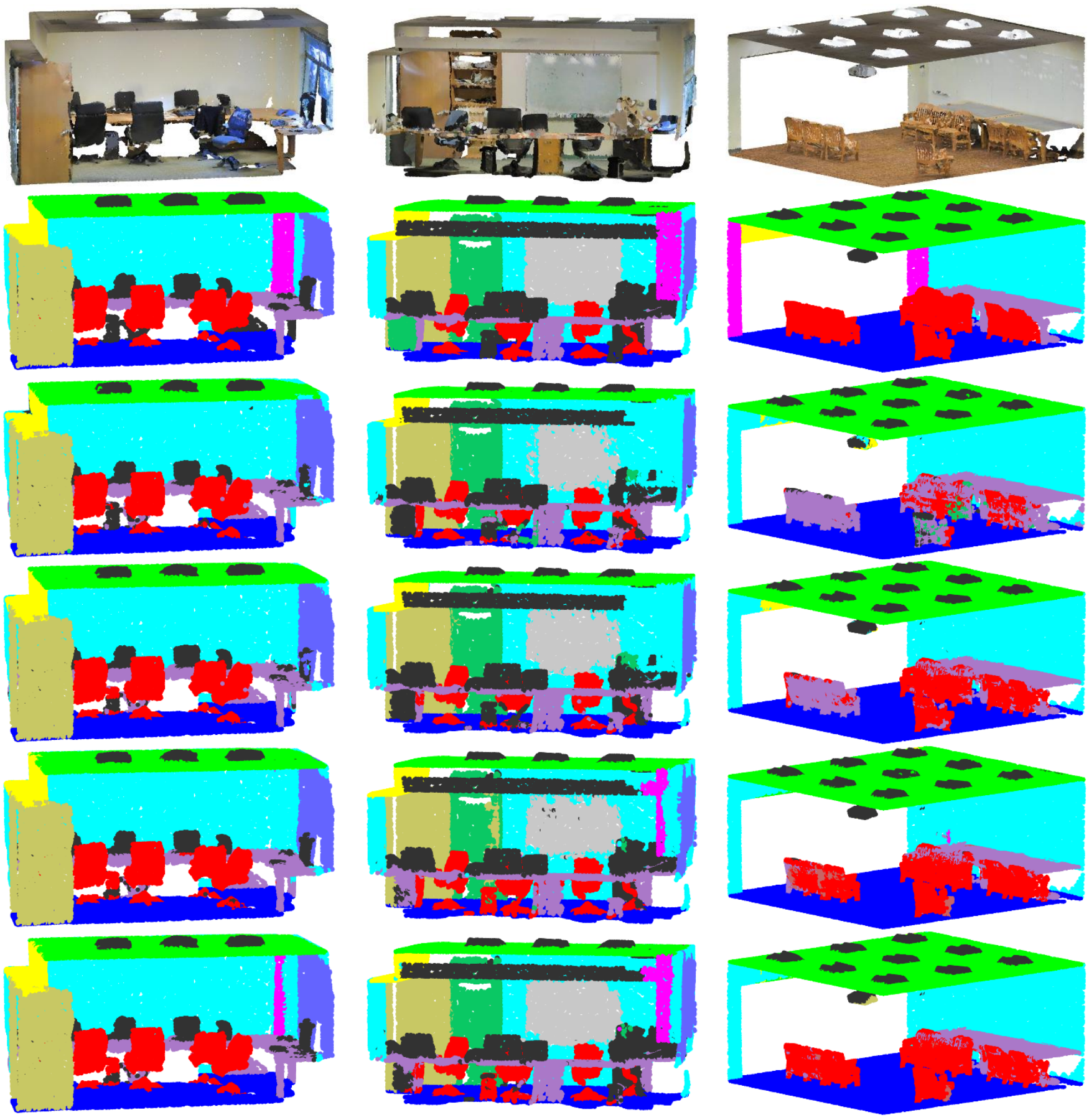}
	\caption{Visualization of S3DIS segmentation results. From top to bottom: input scene, ground truth, baseline, baseline-E, RCNet, RCNet-E.}
	\label{fig:S3DIS}
\end{figure}

\subsubsection{Results} We compare our RCNet with PointNet \cite{PointNet}, A-SCN \cite{ShapeContextNet} and Pointwise CNN \cite{PointWise}. The results are reported in Table~\ref{table:S3DIS}. As is shown, our RCNet improves A-SCN by about $ 0.5\% $ in mean IoU and $ 2\% $ in overall accuracy. We visualize a few segmentation results in Fig.~\ref{fig:S3DIS}. It can be observed that RCNet is able to output smooth predictions and segment the small objects well. In contrast, the baseline methods tend to produce large prediction errors. This shows the benefits of our recurrent set encoder and the 2D CNN as feature aggregators. With ensemble, the segmentation accuracy is further boosted and our RCNet-E achieves the best results.

\subsection{Architecture Analysis}

In this section we show the effects of network hyper-parameters and validate
the design choices through a series of controlled experiments. We consider the following two
main contributory factors on model performance: (1) the size of beams; (2) the number of points. We use ModelNet40 dataset as the test bed for comparisons of different options. Unless explicitly noted, all the experimental settings are the same with those in the shape classification experiment.

\subsubsection{The Size of Beams} The beam size controls how much local context information would be utilized, and is a major contributory factor for the network performance. For RCNet, large beams will lead to a small feature map for the downstream CNN. This would increase the efficiency of CNN but in turn result in the loss of fine-scale geometric details. Moreover, beams with large size would be filled with too many points, and as a result the RNN would perform poorly in feature modeling. On the other hand, if the size of beams is too small, the subregions would contain insufficient amount of points, which is adverse to the feature learning.

We conduct several experiments to investigate the influence of beam size on the network performance. In particular, we test RCNet with different specifications of hyper-parameters $ r $ and $ s $. The results are reported in Table~\ref{table:beam_size}. As is shown, both larger and smaller beam sizes would hurt the performance, and $ r\times s =32 \times 32 $ leads to the best results. Note that, although beam size is an important parameter on the performance, our RCNet is still quite robust to this factor. In contrast, the max-pooling based encoder behaves quite sensitively and the performance decreases a lot with large beams. This further validates that pooling is a relatively coarse technique for exploiting geometric details.

\begin{table}[t!]
	\begin{center}
		\resizebox{1.0\columnwidth}{!}{
		\begin{tabular}{c|c|c|c|c|c}
			\hline
			$ r \times s $ & $ 8 \times 8 $ & $ 16 \times 16 $  & $ 32 \times 32 $ & $ 64 \times 64 $ & $ 128 \times 128 $ \\
			
			\hline
			
			Baseline & 77.2 & 86.3 & 89.1 & 89.3 & 86.7  \\
			RCNet & 87.5 &  90.2 & 91.6 & 90.9 &  89.8 \\
			
			\hline 
		\end{tabular}
		}
		\caption{The influence of beam size on network performance. The smaller the hyper-parameters $ r $ and $ s $, the larger the beams, and vice versa. The experiments are conducted on ModelNet40, and the metric is classification accuracy.}
		\label{table:beam_size}
	\end{center}
\end{table}

\begin{table}[t!]
	\begin{center}
		\resizebox{1.0\columnwidth}{!}{
		\begin{tabular}{ccccc}
			\hline
			\# Point & Baseline + DP & RCNet + DP & Baseline & RCNet  \\
			
			\hline
			
			1024 & 88.9 & 91.1 &  88.2 & 90.2\\
			512 & 88.2 & 90.4 & 68.2 & 76.2\\
			256 & 87.7 & 90.2 & 35.3 & 38.1\\
			128 & 86.4 & 87.8 & 17.8 & 24.9\\
			
			\hline 
		\end{tabular}
		}
		\caption{Experiments on robustness to non-uniform and sparse data. DP stands for random point dropout during training. The experiments are conducted on ModelNet40.}
		\label{table:num_points}
	\end{center}
\end{table}

\subsubsection{The Number of Points} Point clouds obtained from sensors in real world usually suffer from data corruptions, which lead to non-uniform data with varying densities \cite{PointNet++}. To validate the robustness of our model to such situations, we randomly dropout the number of points in testing and conduct two different groups of experiments. In the first group, the models are trained on uniform point clouds without random point dropout, while in the second group the models are trained with random dropout as well. In the experiment, we set $ r=s=32 $ as in the shape classification task. The results are shown in Table~\ref{table:num_points}. We observe that models trained with random point dropout (DP) during training are fairly robust to the sampling density variation, with drop of accuracy less than 3.3\% when point number decreases from 1024 to 128. In contrast, those trained only on uniform data fail to generalize well to the cases of non-uniform data. Note that, despite the drop of accuracy, our RCNet still achieves better performance than the baseline model when trained without DP. This validates the superiority of RNN in subregional feature extraction compared to max-pooling.


\section{Conclusion and Discussion}
\label{sec:conclusion}

In this work we present a new deep neural network for 3D point cloud processing. Our network consists of a recurrent set encoder and a 2D CNN. The recurrent set encoder partitions the input point clouds into several parts, which are encoded via a shared RNN. The encoded part features are later assembled in a structured manner and fed into a 2D CNN for global feature learning. Such design leads to an efficient as well as effective network, thanks to the benefits of CNN and RNN. Experiments on four representative datasets show that our method competes favorably with the state-of-the-arts in terms of accuracy and efficiency. We also conduct extensive experiments to further analyze the network properties, and show that our method is quite robust to several key factors affecting the model performance.

Finally, we note that the proposed recurrent set encoder can be generalized to other contexts. For example, we can build a KNN graph for the input point cloud and model the local neighborhood for each point with recurrent encoder. In particular, we can sort the $ k $ nearest neighbor points according to their distances to the query point, and then apply RNN to this point sequence for local feature learning. This is different from KCNet \cite{Kernel-Graph} which uses a local point-set kernel, and will be explored in the future.


\section{Acknowledgments}
This work was partially supported by NSF IIS-1718802, CCF-1733866, and CCF-1733843.

\bibliographystyle{aaai}

\end{document}